# research

DOI:10.1145/3608473

**Assessing the environmental impacts of machine learning on microcontrollers.**

BY SHVETANK PRAKASH, MATTHEW STEWART, COLBY BANBURY, MARK MAZUMDER, PETE WARDEN, BRIAN PLANCHER, AND VIJAY JANAPA REDDI

# Is TinyML Sustainable?

THE CONTINUED GROWTH of carbon emissions and global waste presents a great concern for our environment, increasing calls for a more sustainable future. In response, the United Nations' (UN) 2030 Agenda for Sustainable Development established a shared framework aiming toward peace and prosperity for people and the planet. At its core are 17 Sustainable Development Goals (SDGs),[48] a call to action for all countries to work toward a more environmentally, economically, and socially sustainable future.

Tiny machine learning (TinyML), which enables ML on microcontroller (MCU) devices, holds potential for addressing numerous UN Sustainable Development Goals, particularly those related to environmental sustainability (see Figure 1). While TinyML's operational benefits for sustainability are often highlighted, it is crucial to consider the entire life cycle of both applications and hardware to ensure a net carbon reduction. This article contributes by presenting case studies illustrating TinyML's sustainability benefits, examining the environmental impacts of TinyML at both MCU and system levels through a life cycle analysis (LCA), and identifying future research directions for sustainable TinyML.

TinyML is the deployment of machine learning (ML) algorithms onto low-cost, low-power, and resource-constrained MCU systems. TinyML stores neural network models directly within memory (for example, flash) and runs inference directly on the output of onboard sensors. This approach enables intelligent on-device sensor analytics unavailable with traditional Internet of Things (IoT) approaches, which instead typically rely on communication with the cloud to transmit data for external processing. Importantly, TinyML achieves this using a fraction of the compute resources needed for traditional ML systems. Table 1 compares TinyML with traditional BigML (such as cloud and mobile systems) and shows how TinyML requires orders of magnitude fewer resources across compute, memory, storage, power, and cost. Finally, while the heterogeneity and limited resources of MCU devices present new challenges for on-device training, model updating, and deployment, recent research, and the development of ML frameworks such as TensorFlow Lite for Microcontrollers[14] have increased the accessibility of TinyML.

» **key insights**

- TinyML (machine learning on low-power microcontrollers) unlocks sustainable computing solutions, increasing agriculture yield and mitigating climate change, to help address many of the UN's Sustainable Development Goals

- Life cycle analysis (LCA) reveals a significant carbon footprint for TinyML at scale, as there are billions of MCUs deployed globally; however, TinyML can drastically reduce emissions in other sectors, offsetting its own footprint in the process

- The microcontroller unit (MCU) in TinyML devices has a relatively small footprint compared to the battery and sensing components; hence, TinyML devices environmental influence rests on holistic and sustainable system design.

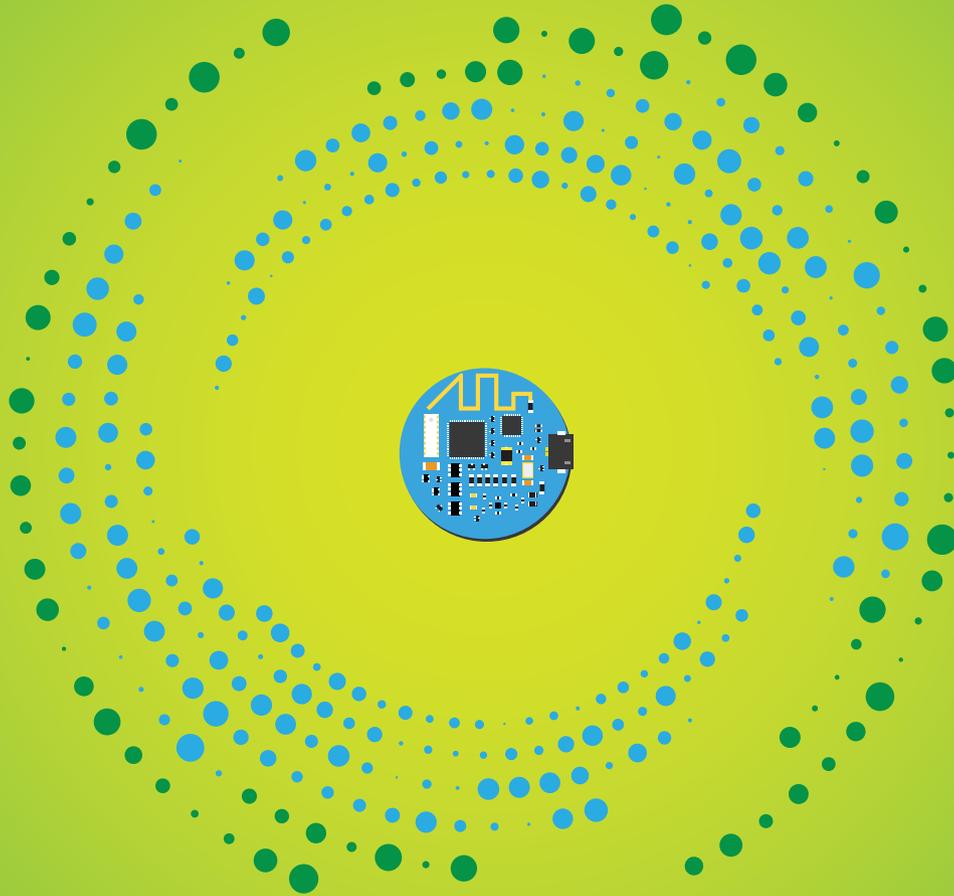

The ubiquity, low-cost, and small power envelope of MCUs, paired with TinyML's independence from Internet connectivity, enables ML models to be deployed globally anywhere at scale. For these reasons, along with bandwidth, latency, energy, reliability, and privacy concerns, running ML directly on these embedded edge devices is growing in popularity.

With more than 250 billion MCUs deployed globally today, and the cost of MCUs expected to drop below $0.50 per unit, this number is expected to grow, eclipsing 40 billion MCUs shipped per annum in the next decade.[39] As such, TinyML will become an ever-present technology. But the question we must ask ourselves is do we run the risk of producing an Internet of Trash over the course of TinyML devices' lifetime?

### Applications of TinyML for Sustainability

To fairly evaluate the environmental impacts of machine learning on MCUs, we first consider TinyML's benefits. Typical well-known consumer-facing applications of TinyML include keyword spotting, image classification, and anomaly detection.[7] However, many other applications of TinyML can be used to enable a more sustainable future.[6] In the following sections, emerging applications are highlighted, which show how TinyML can help with important environmental-related SDGs (as shown in Figure 1).

TinyML is well-suited for improving the sustainability of global agriculture, aiding wildlife conservation, and helping combat climate change and its impacts.

**Zero hunger and good health and well-being (SDG #2 and #3).** *End hunger, achieve food security & improved nutrition, promote sustainable agriculture, and ensure healthy lives while promoting well-being for all at all ages.*

ML applications can increase agriculture production through data-driven methods. For example, Nuru, a mobile and cloud-based ML app from the PlantVillage project, is more accurate than humans at detecting plant diseases and enabled one farmer to increase her revenue by 55% and yields by 146%.[12,30] ML has also been used for autonomous devices such as tiny drones, which can provide targeted pesticide applications that reduce use to 0.1% of conventional blanket spraying.[27] As another example, researchers developed a cough monitor system to flag respiratory problems in pigs by placing microphones over animal pens that can alert farmers 12 days earlier than standard methods.[27]

TinyML has the potential to increase the impact of these systems. First, it can enable these and many other applications to be used in remote regions through low-power, low-connectivity operations. Second, it can enable scaled deployment of these smart sensors, which could provide more targeted information (for example, on all individual pigs in real time for the cough monitor system). Most importantly, it would increase global access to these technologies by reducing costs. As Sparrow and Howard note, global adoption can only occur if devices "can be manufactured and sold cheaply enough to be available to smaller farms."[41]

TinyML can also be used to aid in our health and well-being. One of the

NOVEMBER 2023 | VOL. 66 | NO. 11 | COMMUNICATIONS OF THE ACM    69

diseases noted in the UN's SDG report is malaria due to its massive global impact. In fact, nearly half of the world population has been killed by mosquitoes.[50] Gaps in funding and access to life-saving tools led to a disproportionate 94% of all malaria cases and deaths in 2019 occurring in the African region.[48] Using Edge Impulse,[22] a development platform for TinyML, a system was prototyped to identify the deadliest mosquitoes using wing beats sound classification with 88.3% accuracy.[47] This is another example in which global access to these systems will have a tremendous impact and could potentially save lives.

**Life on land and below water (SDG #14 and #15).** *Protect, restore, conserve, and promote sustainable use of terrestrial and aquatic ecosystems, sustainably manage forests and marine resources, combat desertification, halt biodiversity loss.*

TinyML can help preserve the planet's biodiversity by improving the efficiency of conservation efforts that rely on distributed sensing networks. One such instance is to resolve human-elephant conflicts in Asia and Africa. By only transmitting notifications of elephant detection instead of full video streams to the cloud, RESOLVE's Wild-Eyes AI camera can run for more than 1.5 years on a single Lithium-Ion battery.[16] AI on the edge is also being used at Liwonde National Park in Malawi to prevent poaching, and as of September 2019, the park had lost no animals in 30 months.[46] Similar systems are being used to prevent collisions with whales in busy waterways. For example, Google deployed a TinyML model on hydrophones (underwater microphones) to alert ships in Vancouver Bay.[24]

Due to the low computational requirements, opportunities also exist for upcycled and recycled electronic devices for TinyML applications. Rainforest Connection (RFCx) uses recycled smartphones to develop solar-powered listening devices for pinpointing deforestation over long distances.[35] Similar opportunities exist for upcycling MCUs.

**Climate action (SDG #13).** *Take urgent action to combat climate change and its impacts.*

TinyML is well-suited to efforts aimed at combating climate change and its impacts through environmental monitoring applications. For example, Ribbit Network recently launched an effort to crowdsource the world's largest greenhouse gas emissions dataset through distributed intelligent sensors that enabled cheap, accurate, and actionable local data on emissions. Similarly, the SmartForest project utilizes a remote monitoring system to provide information on tree growth. This replaced the need for 150–160 employees to regularly go into the field with a single trip to install the sensors,[15] significantly reducing human impact on the ecosystem while increasing data quality.

In the long term, TinyML also has the potential to power the next generation of tiny robots to help reduce the global impact of climate change. For example, climate change has contributed to the widespread decline of essential pollinators like bumble bees,[40] threatening the global food supply (SDG #2 mentioned previously). TinyML can help provide intelligence to tiny robots like the Robobee[51] that can be used as artificial pollinators. However, there are still many challenges and opportunities to unlock tiny robot learning.[31]

Finally, one area of broad interest is the building sector. Existing systems that control lighting, automated window shading, and HVAC based on occupancy and light-intensity sensors show a 20%–40% reduction in building energy usage.[5,28] Adding ML capabilities to these systems would lead to further improvements in efficiency. This increased efficiency is critical as energy production along with residential and commercial energy usage are leading sectors contributing to global greenhouse gas emissions (see Figure 2).

## Quantifying the Sustainability of TinyML

The benefits of ML on MCUs for environmental sustainability and beyond will continue to fuel the IoT revolution, connecting billions of devices around us. However, embedding smart computing into everyday objects may have looming environmental consequences through increased electronic waste.[21] To better understand the environmental costs associated with TinyML, a life cycle analysis (LCA) of the complete TinyML system (that is, MCU plus peripherals and power supply) is performed. This analysis demonstrates the footprint of MCUs and TinyML systems individually is relatively small. When this analysis is expanded to consider the global scaled impact of TinyML, the impact could be substantial if not offset by using TinyML for sustainable applications.

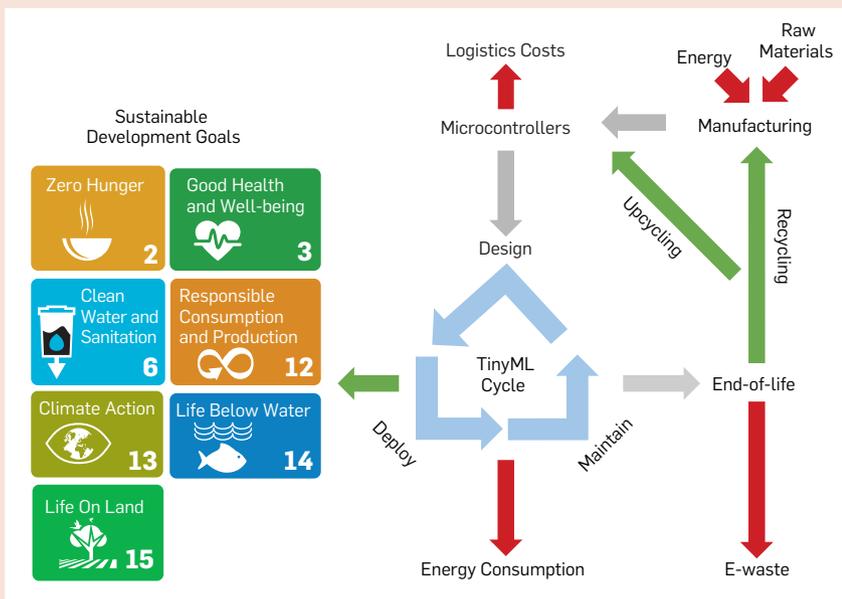

**Figure 1.** We show the positive (green arrows) and negative (red arrows) environmental footprint of the complete life cycle of TinyML systems as well as how TinyML can contribute to the UN's environmental sustainability goals.

**Table 1.** Cloud and mobile ML systems compared with TinyML across frequency, memory, storage, power, price, and footprint. The footprint of TinyML systems is far less.

| Platform | Freq. | Memory | Storage | Power | Price | $CO_2$-eq Footprint |
| --- | --- | --- | --- | --- | --- | --- |
| Cloud | GHz | 10+GB | TBs-PBs | ~1 kW | $1000+ | Hundreds of kgs |
| Mobile | GHz | Few GB | GBs | ~1 W | $100+ | Tens of kgs |
| **Tiny** | **MHz** | **KBs** | **Few MB** | **~1 mW** | **$10** | **Single kgs** |

**Growing environmental risks of IoT trash.** Electronic waste (e-waste) is a growing concern and polluting our environment. In 2019, it was reported that e-waste had grown by 20% over the past five years,[48] and by 2030, forecasts predict a total of 75 million metric tons of e-waste.[43] In addition to the e-waste, the carbon emissions from manufacturing and operating these devices are also growing and impacting the environment. TinyML has the potential to drive more demand for innovative IoT solutions that would advance the ubiquitous computing movement, but further exacerbate the growing "Internet of Trash."

Parallels can be drawn from the plastic pandemic. An abundance of resources (for example, plastic, silicon) has made it easy to manufacture "infinitely" at scale and the convenience offered by such products made it easy to ignore environmental concerns. Consequently, plastic has contributed significantly to land and water pollution, and its production contributes to global warming by emitting greenhouse gases. Plastic also contains toxic chemicals that can leach into food and water and have been linked to health problems. These adverse side effects provide a cautionary tale and motivation to carefully consider the net benefit of TinyML systems and applications.

**Environmental impact of MCUs.** The TinyML life cycle analysis starts at the MCU level with publicly accessible data from STMicroelectronics.[45,a]

a  The general trends hold for other MCU manufacturers.

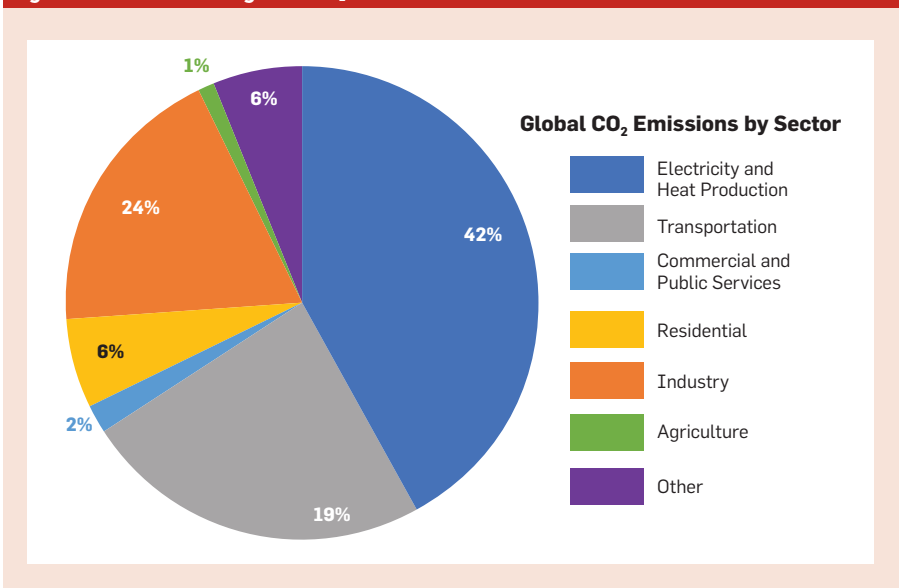

Figure 2. Breakdown of global $CO_2$ emissions as of 2019.[23]

Global $CO_2$ Emissions by Sector
- 42% Electricity and Heat Production
- 19% Transportation
- 2% Commercial and Public Services
- 6% Residential
- 24% Industry
- 1% Agriculture
- 6% Other

The hardware life cycle of an MCU can typically be broken down into five stages: extraction and treatment of raw materials, product manufacturing, transport and distribution, product use, and end of life. As shown in Figure 3, there are four different environmental indicators that can be used to analyze the footprint of the processing hardware required for TinyML: water demand, freshwater eutrophication, photochemical oxidant formation, and climate change. Across all four indicators, production, or more specifically, energy consumption during production, is the dominant driver of an MCU's environmental footprint, as noted in previous work.[25,52] However, the exact breakdown varies across indicators.

*Water demand.* SDG #6 highlighted that billions of people are without abundant access to clean water. This indicator measures the volume of water evaporated, consumed, used for cooling, or released downstream, during an MCU's life cycle. Figure 3 shows that while much (54%) of the water used in an MCU's life cycle is attributed to the production site, extracting and transforming the raw materials also requires a substantial amount of water (41%).

*Freshwater eutrophication.* Eutrophication, the proliferation of algae and plants in bodies of water, is one of the most significant threats to our aquatic ecosystems (SDG #14). This indicator measures this impact in grams of phosphorous equivalent (g P-eq.), as phosphorous is a common cause of algae blooms from over-enriched aquatic ecosystems. Figure 3 shows that this environmental indicator has the most balanced impact on the five stages of an MCU life cycle, with 45% of the footprint attributable to production, 28% to the MCU use, and 27% to the extraction of raw materials.

*Photochemical oxidant formation.* This indicator measures milligrams of non-methane volatile organic compounds (NMVOC) formed. These play an essential role in the formation of photochemical oxidants, which can exacerbate respiratory ailments and lead to smog formation, impacting the climate (SDG #13), local air quality (SDG #15), and public health. Figure 3 shows that this footprint is mainly driven by production, accounting for 74% of the total, with 71% coming from energy usage during production.

*Climate change.* This indicator measures equivalent grams of carbon dioxide ($CO_2$-eq) emitted. $CO_2$ is the most prevalent greenhouse gas produced by humans and a primary driver of climate change (SDG #13). As Figure 3 shows, most of the carbon emissions come during production of the MCU (81%), with the majority resulting from energy consumption (56%). The entire carbon footprint of an MCU is 390g $CO_2$-eq. For perspective, this footprint is equivalent to a gasoline-powered car driving 1.6km. Given that cars typically drive hundreds of thousands of miles during their lifetime, a single MCU alone has minimal impact in the context of everyday human actions. In the following section, $CO_2$ emissions are used as the primary measure due to their wide acceptance for assessing environmental impact.

**Footprint of TinyML systems.** MCUs are the heart of embedded TinyML systems, but we must consider the additional components that constitute a complete TinyML system to get a more accurate picture of the complete footprint. Thus, in this section, we systematically analyze the footprint of the systems used for the widely deployed TinyML applications of keyword spotting and image classification. This analysis outlines all pieces needed for deploying a system in the wild such as casing, sensing, actuators, transport, and more.

*TinyML Footprint Calculator.* We developed an open source TinyML Footprint Calculator to evaluate the footprint of TinyML systems.[b] This tool can be used in the future to help engineers understand the impact of the devices they are developing. For example, this tool could produce the environmental impact report for ML sensor datasheets.[49]

Our calculator leverages the raw data from a recent 2021 study by Pirson and Bol[32] assessing the embodied carbon footprint of IoT devices. Pirson and Bol[32] break down the general architecture and hardware profile of an IoT edge device into a collection of basic functional blocks: processing, memory, actuators, casing, connectivity, PCB, power supply, security, sensing, transport, user interface, and others circuit components (for example, resistors, capacitors, diodes).

Within the various blocks, Pirson

---

b Available at https://github.com/harvard-edge/TinyML-Footprint

and Bol break down the impact based on application specifications. However, Pirson and Bol note that the data provided only encapsulates stages 1 through 3 of the hardware life cycle (that is, embodied footprint). As such, we additionally model and capture the product use stage (operational footprint) and end-of-life stage of the hardware life cycle in Figure 4.

To account for the use stage, we calculated the $CO_2$-eq of recharging a power supply for three years of continuous use at 1mW, the average power consumption of commercial TinyML systems.[7,29] We note that some TinyML applications may require much less power when idle (for example, keyword spotting) and that we used a three-year benchmark to be consistent with Apple's analysis that was our baseline.

In addition, we included the ML model training costs since they can be large, on the order of millions of kg

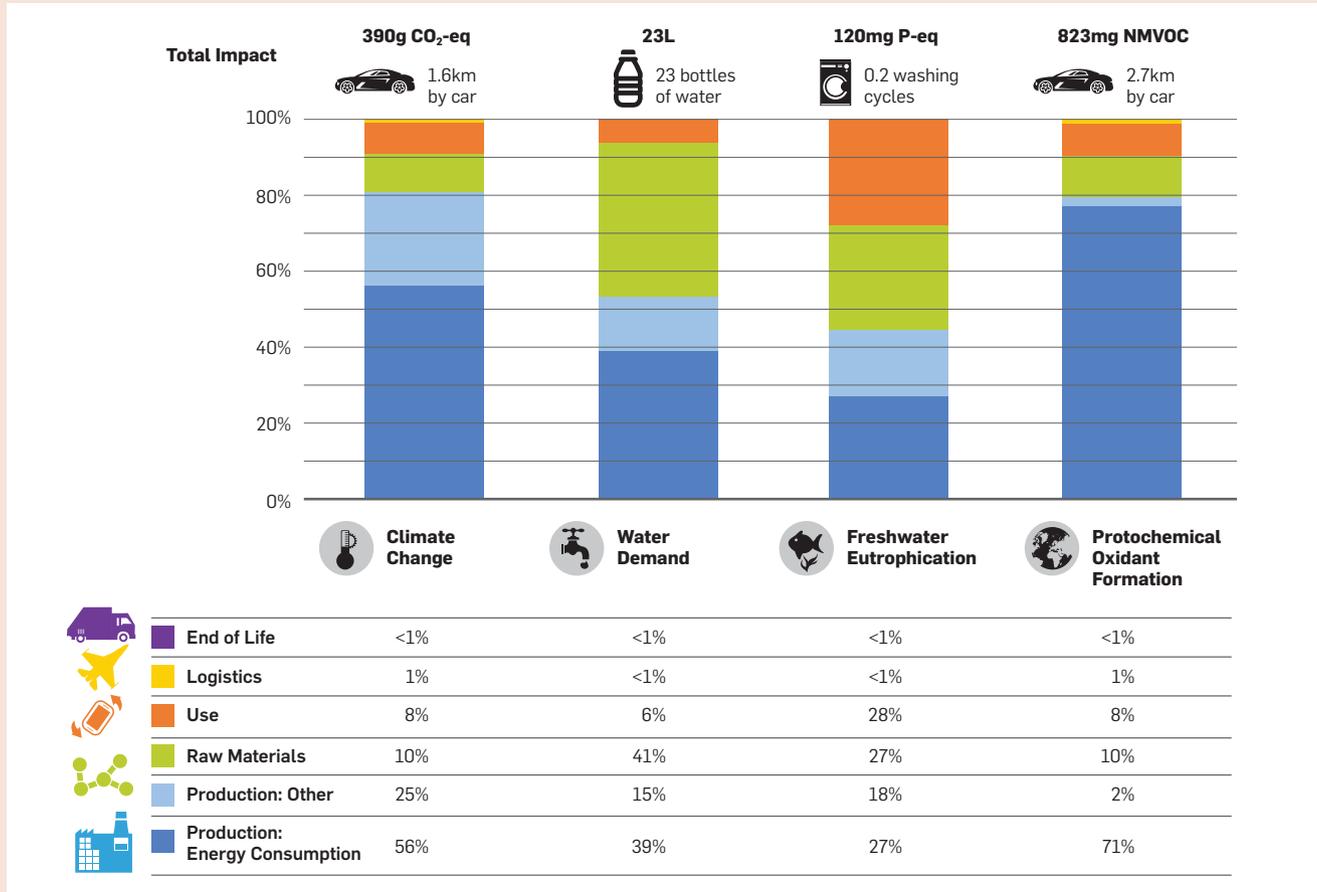

Figure 3. Four different environmental indicators measuring the impact of MCUs on our environment.

Each footprint contains both the operational and embodied footprint of the device, including the five-stage life cycle of an MCU. Data courtesy of STMicroelectronics.[45] The data from other MCU providers follow the same operational and embodied footprint trends.

| Total Impact | 390g $CO_2$-eq<br>1.6km by car | 23L<br>23 bottles of water | 120mg P-eq<br>0.2 washing cycles | 823mg NMVOC<br>2.7km by car |
|---|---|---|---|---|
| | Climate Change | Water Demand | Freshwater Eutrophication | Protochemical Oxidant Formation |
| End of Life | <1% | <1% | <1% | <1% |
| Logistics | 1% | <1% | <1% | 1% |
| Use | 8% | 6% | 28% | 8% |
| Raw Materials | 10% | 41% | 27% | 10% |
| Production: Other | 25% | 15% | 18% | 2% |
| Production: Energy Consumption | 56% | 39% | 27% | 71% |

of $CO_2$-eq for large cloud ML tasks.[52] Costs were based on footprint estimates of DenseNet,[17] which serve as an upper bound on computation as it is an order of magnitude larger than the typical size of TinyML models. This upper bound can also help account for extra training runs that may be conducted during neural architecture search (NAS).[c]

*Breaking down TinyML's footprint.* The calculated footprint of TinyML systems is broken down into three scenarios. The "Low-Cost Profile" scenario represents a keyword spotting application that requires only a simple microphone sensor. The "Medium-Cost Profile" scenario represents an image classification application that requires a much larger camera sensor.

---

c   NAS has been used in prior work,[6] but is not typical for many TinyML application and can be generalized.[11]

The "High-Cost Profile" scenario again uses the image classification application, instead using the upper bound carbon emission values for each component provided in Pirson and Bol.[32] These scenarios represent typical, but not absolute, bounds for assessing classical TinyML systems. For complete details, see https://github.com/harvard-edge/TinyML-Footprint.

As the stacked bar graph on the right side of Figure 4 shows, the embodied footprint of all components is much greater than the system's operational footprint (captured in "Product Use"). This result aligns with previous literature suggesting that manufacturing dominates the environmental footprint of small electronics.[25,52] Moreover, the figure highlights that the processing's (that is, MCU's) embodied footprint does not contribute significantly. Instead, most of the footprint is attributable to the embodied footprint of the additional components (for example, power supply, sensor, circuit board) and the transportation costs associated with manufacturing and distribution. In particular, the power supply is one of the dominating factors in the footprint. The embodied footprint of a battery required to deploy TinyML in the wild for years is much larger than any other system component.

One may consider comparing our TinyML system footprint to another device used for making progress toward the SDGs or an edge-class server. However, the documentation of such data is still relatively new and limited. Thus, using what is publicly available, the typical TinyML system footprint is compared with the latest Apple Watch Series 7 (representative of an "edge" device) to provide a baseline reference for understanding the total carbon footprint of a TinyML device, as shown

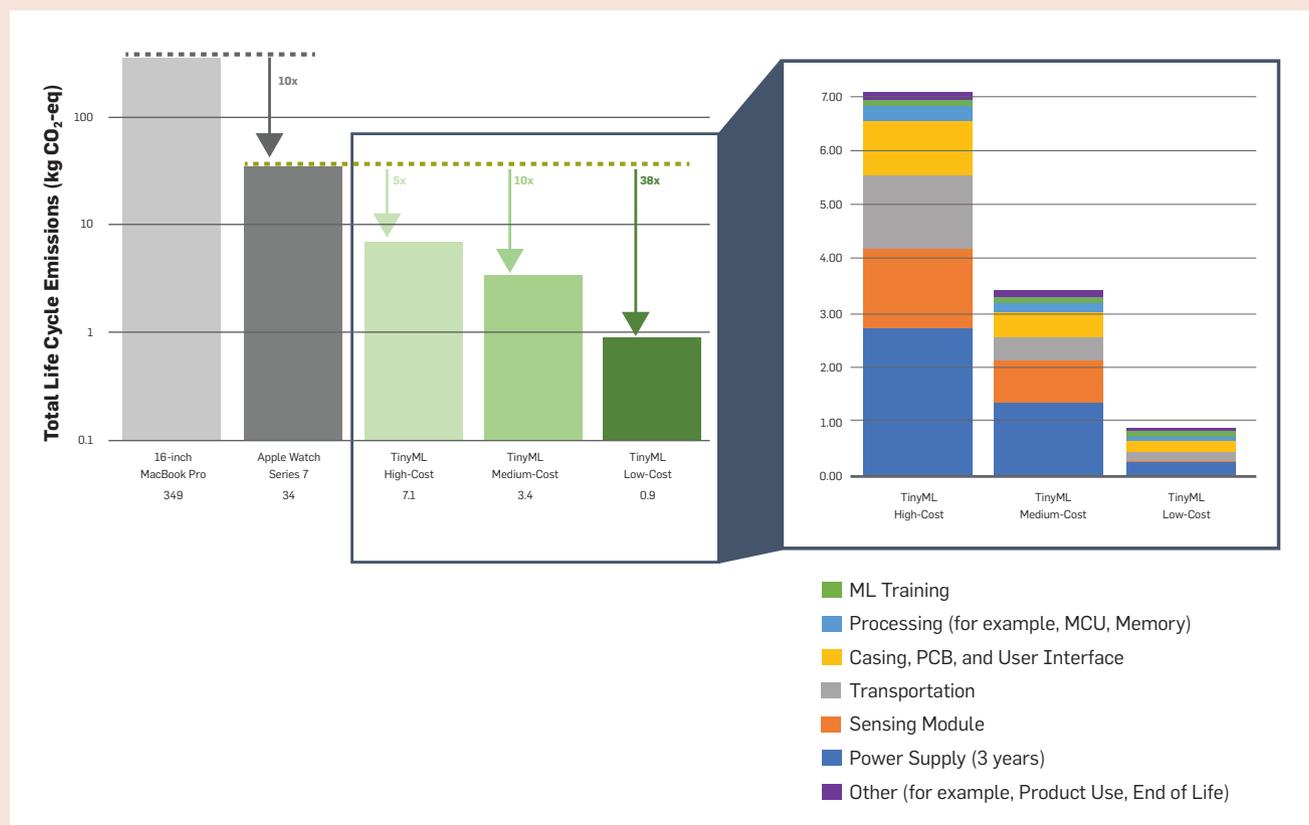

**Figure 4. A breakdown of different TinyML system footprints.**

This illustration highlights the footprint is largely attributable to the embodied footprint of the power supply, onboard sensors, and transportation. Note that actuator and connectivity blocks from Pirson and Bol[32] are encapsulated in "Other" and "Processing," respectively, while "Product Use" captures the operational footprint. The carbon footprint of Apple's Series 7 Watch[4] and 16-inch MacBook Pro[3] are also provided for reference. For more details and to compute the footprint of your own TinyML system, see https://github.com/harvard-edge/TinyML-Footprint.

on the left of Figure 4. In the figure, the footprint of a 16-inch MacBook Pro is also provided to give the reader an idea of a device footprint representative of traditional computing hardware.

The carbon footprint of the Apple Watch, considering three years of use, is 34kg $CO_2$, with 76% of the footprint attributable to production, 10% to transport, and 13% to everyday use.[4] This carbon footprint is 5-38× larger compared to a TinyML system. Moreover, for reference, a TinyML system has a 49-392× smaller footprint than a Macbook.[3]

This complete LCA shows the additional components that constitute a TinyML system have a larger carbon footprint than the MCU alone, but still have a much smaller footprint than existing computing systems.

**TinyML at scale.** Each TinyML device will have an associated environmental footprint, as outlined previously, and can provide environmental benefits. However, humanity is on the path to a future with billions of deployed intelligent IoT devices. To better understand the net effect of TinyML at scale, this section assesses what happens to TinyML's footprint if these systems are scaled to the number of MCUs deployed globally, which currently sits at around 250 billion. For reference, there are currently only around 15 billion IoT devices,[42] with two billion TinyML device installs in 2022 and a projected 11 billion in 2027.[2] These numbers suggest that including all 250 billion existing MCUs under TinyML's footprint is a very conservative overestimate. However, we do so, and use the "High-cost Profile" data from Figure 4, to provide an upper bound for the net effect of TinyML both now and into the near future. This scenario results in a combined, non-trivial global carbon footprint of 1,765 million metric tons of $CO_2$-eq.

This is a substantial footprint that is quite concerning for TinyML on its own. However, along with this number we need to consider the emissions that were avoided by using these systems. As mentioned earlier, there are existing examples (for example, Asopa et al.[5] and King et al.[28]) of simple, intelligent IoT devices that can reduce building $CO_2$ emissions by at least 20%. In this case, enduring the footprint of these smart devices is worthwhile as their footprint is most likely negligible when compared to the 20% reduction in a building's emissions.

Figure 5 now compares the calculated global carbon footprint of TinyML (blue bar) in the context of the emissions these TinyML systems could help avoid through efficiency improvements in other sectors. If the 20% reduction in a single building's emissions were applied to the entire residential home sector over three years (green bar), 1,181 million metric tons of $CO_2$-eq would be avoided. These avoided emissions alone would offset 67% of the worst-case costs of TinyML. The residential sector, though, only represents 6% of total global emissions (Figure 2). If remaining TinyML devices were able to reduce emissions from all other sectors by as little as 0.6% on average (orange bar), then TinyML would break even from an emissions standpoint.

Furthermore, if we were to extrapolate this 20% reduction in the residential sector to all sectors (yellow bar) we would see a net reduction in global $CO_2$ emissions by over 18.4 billion metric tons. While a 20% reduction in emissions from all sectors may be unrealistic, anything greater than a 0.6% reduction on average would result in TinyML saving more emissions than it produced. This result suggests that TinyML systems, if designed with careful intention, can elicit an overall positive impact on the environment.

### Discussion

Prior claims regarding the use of digital technologies for greenhouse gas emissions mitigation do not always address critical aspects that can result in overestimated benefits.[36] Thus, in this section, we recognize the limitations of our analysis and discuss other important factors that should be considered in future work to develop a deeper understanding of the impact of TinyML on environmental sustainability.

**Limitations of our study.** One major limitation is the lack of publicly available data on the environmental impact of modern digital electronics. This makes it difficult for our analysis to be detailed and precise and makes it challenging for consumers to understand the environmental impact of their purchases. This can also lead to uncertainty in analysis that can be challenging to quantify. However, it is promising to see there is increasing demand for LCA and carbon footprint data. This additional data will empower consumers and enable the heterogeneity in these systems to be accounted for in future work.

Another important consideration is Jevons' paradox (or rebound effect), which suggests that advancements in efficiency can lead to an overall increase in consumption and a negative impact on the environment. The section on "TinyML at scale" attempts to address this by examining a scenario in which TinyML systems are produced at a much larger scale than they are used today. And while this section suggests that TinyML systems have the potential to have a positive impact, the conclusions of this assessment are limited

**Figure 5. If all 250 billion MCUs were TinyML systems with three-year lifespans, their worst-case footprint would be 1,765 million metric tons of $CO_2$.**

If these systems enabled a 20% emissions reduction for the residential sector and only a 0.6% reduction for all other sectors (Figure 2), the total footprint would be net-zero. Anything larger (for example, 20%) results in more carbon savings from TinyML than emissions.

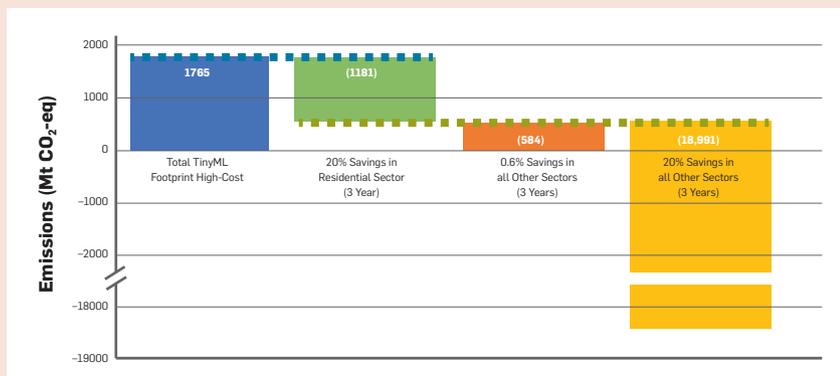

by the current availability of heterogeneous data.

Finally, it is also important to note that our analysis approximates carbon savings from TinyML solutions by comparing with a baseline with no intervention. However, alternative (non-TinyML) approaches could also be made to save emissions (for example, behavioral changes or manual efforts by humans to reduce building emissions) that future works should compare with to provide further trade-offs and analysis.

**Considerations for future studies.** In this section, we mention several additional factors that should be considered in future complementary studies.

*IoT device growth.* Our study assumes all 250 billion existing MCUs are TinyML systems as a conservative overestimate for analysis. However, this is not actually the case as only a subset of deployed MCUs are being used for TinyML. Realistically, 250 billion TinyML systems will not be reached for some time, though. Table 2 projects when this 250 billion device mark will be reached based on available IoT device reports from Statista.[42] IoT device growth can be closely tied to TinyML's growth.[1] Assuming linear growth, the 250 billion device mark will be reached in 120 years. However, if we instead assume exponential growth, the mark would be reached in only 20 years, and we would reach trillions of devices in 30 years.

Given this, a key question is whether there is a limit to the carbon savings from TinyML, leading to another example of Jevons' paradox. Additionally, while most commodity MCUs can support TinyML applications today, a phase-in process will likely be necessary to transition current IoT devices.

*Semiconductor manufacturing.* Using data from Pirson and Bol, our analysis assumes the MCU is fabricated using 90nm CMOS technology.[32] However, the environmental impact of semiconductor manufacturing increases with each successive technology node. For example, Bardon et al.[9] shows a 2.5× increase in greenhouse gas emissions per wafer when scaling from 28nm to 3nm. Fortunately, emissions are reduced when normalized on a per-transistor basis. Therefore, if we are to keep our MCU designs the same (that is, the same number of logic and transistors) then our MCU footprint can actually reduce and benefit from each successive technology node.[d] We additionally advocate for repurposing existing MCU hardware for TinyML systems.

**Table 2. Exponential and linear growth projections based on Statista Reports for IoT devices.[42] Each column lists the first year that *X* billion devices will be reached.**

| | IoT Device Growth | | | | |
|---|---|---|---|---|---|
| | ~15 Billion | >50 Billion | >100 Billion | >250 Billion | >1 Trillion |
| Linear | 2023 | 2041 | 2067 | 2144 | 2531 |
| Exponential | 2023 | 2032 | 2036 | 2043 | 2053 |

*Device lifetime: Embodied vs. operational footprint.* Our study assumes a three-year device lifetime to compare with LCA's from other vendors. However, Eeckhout highlights that the ratio of operational to embodied footprint is important.[18] In the case of our study, sweeping the device lifetime from 1–10 years does not have a large impact on the carbon footprint of an individual device as the operational footprint is minimal.[e] However, when considering the saved emissions from TinyML, the lifetime of the devices impacts the total positive impact. Reducing our assumed three-year lifespan to one year would require 4.1% savings from other sectors to break even (rather than 0.6%). However, increasing the lifespan to 10 years would allow the residential sector savings alone to offset the total footprint of TinyML.

*Beyond carbon.* Our study concludes that TinyML devices can elicit an overall positive impact on the environment with respect to carbon emission savings and global warming.

However, it is important to note that there are many other environmental factors to consider. As such, it is possible that TinyML does not have an overall positive impact when considering other environmental measures for sustainability. Finally, climate change represents a significant planetary boundary, but it is essential to acknowledge the existence of other environmental factors within which humanity must safely operate. These include but are not limited to biosphere integrity, global change

---

d  This benefit would also apply to the image sensor used in our study that assumes fabrication using 250nm CMOS technology.

e  However, if the power consumption of the TinyML system is much greater and requires multiple batteries, the embodied footprint would increase.

---

biology, hydrological and biogeochemical cycles, land-system changes, stratospheric ozone depletion, and ocean acidification.[44]

Another essential aspect to consider is the societal and human costs associated with TinyML production. TinyML deployment is inherently different from cloud ML deployment, meaning TinyML's non-environmental related impacts could also look very different. For example, batteries are a critical component of TinyML systems, the implications of which are discussed later. Moreover, the global and remote distribution of TinyML systems means that the associated costs are not localized to a single location, contrasting a cloud datacenter. Thus, rules and regulations surrounding best practices for manufacturing and deploying TinyML systems can be subject to exploitation as laws regarding mining, child labor, and others can change based on location. We must also consider these consequences.

**Future Sustainable TinyML**

While TinyML has the potential to contribute to global sustainable development and environmental sustainability, there are still many challenges that must be overcome to fully realize this potential. As noted, the environmental impact of TinyML will be non-negligible. Even if the benefits of TinyML can potentially outweigh its impact, it is important to be cautious and ensure future generations of TinyML devices are sustainable. In this section, we will discuss the broader implications of our study and suggest ways to make TinyML more sustainable.

First and foremost, it should be noted that for sustainable design to be truly successful, incentives must be provided to corporations and engineers to prioritize sustainability when making decisions. Policy changes such as a carbon tax or related measures that can help curb emissions need to be investigated for consideration. Moreover, as mentioned, human cost needs to be also pushed at the forefront of any decisions or innovations made.

*Energy harvesting.* Our analysis revealed that the batteries used to power TinyML devices dominate their environmental impact. Batteries also present several other environmental issues, such as pollution and the release of carcinogens.[26] In particular, the extraction of lithium is especially harmful to the environment and has caused severe water insecurity for already marginalized people.[19]

Research in energy harvesting[38] needs to be prioritized to make "batteryless" TinyML the standard practice to reduce the associated environmental and societal costs that come with batteries. Furthermore, advancements in intermittent computing[20] could be suitable in TinyML scenarios and help further reduce the needed power supply.

*Efficient sensing.* The second largest contributor to the footprint in two of the three profiles analyzed was the sensor. Sensing is essential in TinyML, but using smaller (for example, camera vs. inertial measurement unit) or lower-quality sensors (for example, low- vs. high-resolution camera) could further reduce the environmental impact of the device. Due to the relatively low footprint of the compute, more advanced TinyML models could be used to make up for the loss in performance introduced by lower-quality sensors, potentially reducing the overall footprint while achieving the same performance. Additionally, sensor fusion, using multiple small sensors, could also be used to reduce the total environmental impact.[13] 

*Datasheets for ML sensors.* Greater transparency regarding the system's data and costs is needed to deploy these TinyML devices safely and ethically. TinyML instantiations must clearly and transparently articulate their privacy and security boundaries. One solution to address privacy concerns is to separate the input sensor data and ML processing from the rest of the system at the hardware level. Also, new supplementary information is needed in the form of a datasheet that builds upon traditional datasheets used for electri-

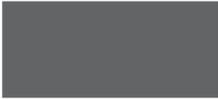

> For sustainable design to be truly successful, incentives must be provided to corporations and engineers to prioritize sustainability when making decisions.

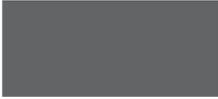

cal components to enable transparency to end users.[49] These datasheets should include information about the environmental impact and LCA of the device in an easy-to-understand format so that users can use TinyML devices in an environmentally friendly and sustainable way. To drive this, a benchmark is needed to systematically report and measure the sustainability of computing systems.

*Datasets for low-resource domains.* Many TinyML applications depend on real-world data, which can be challenging to obtain, particularly in public domains. To foster TinyML development, there is a need for extensive, open access datasets focused on low-resource, high-impact sensor-based problems, akin to ImageNet for TinyML. Similar to how open source software development has enabled code sharing and reduced costs, creating large, public, and representative datasets for TinyML is essential. However, it is also important to consider the environmental impact of the data collection. For example, collecting data in nature can be disruptive and harmful. On the other hand, alternative methods such as computationally intensive simulations may require carbon-consuming resources to obtain.

*Emerging technologies.* New technologies are being developed that could lead to more sustainable TinyML practices. One example includes flexible electronics: PragmatIC Semiconductor has reported less than half of a single gram of $CO_2$-eq manufacturing such integrated circuits.[33] This class of technologies could enable TinyML to achieve greater reductions in emissions than anticipated in our analysis. However, these technologies are not yet mature and have less processing power than traditional silicon devices.[10] More research and development is needed to utilize these sustainable technologies fully. Furthermore, sustainability needs to be made a measure of fab performance for existing technologies as there are many opportunities for potential energy savings in which TinyML solutions could help.[34]

*Recycle and upcycle.* TinyML can potentially exacerbate the problem of electronic waste. However, recycling and reusing TinyML devices is a viable option as many of the algorithms can run on standard, commonly used MCU hardware. This can extend the life of the MCU and reduce the amount of waste sent to landfills. In addition to recycling MCU hardware, industry should consider recycling old technology nodes when fabricating additional MCUs for TinyML. More broadly speaking, TinyML systems should consider more sustainable designs when high computational performance is not an application requirement.

*Accessibility.* Finally, for TinyML to have a significant impact on a global scale, there is a need for global access to hardware and educational resources. Fortunately, recent efforts, led by the TinyML foundation and the TinyML Open Education Initiative (TinyMLedu),[f] among others, have both developed such open-source materials and provided low-cost or no-cost hardware to learners.[37]

## Conclusion

Using ML on microcontrollers can have a significant impact on environmental sustainability. Low-power ML on low-cost MCU-class hardware has the potential to improve efficiency in various sectors, enabling significant reductions in carbon emissions. This assessment shows that TinyML's carbon footprint could be offset by using the technology to reduce emissions from other economic sectors. However, TinyML's footprint is not negligible when scaled globally, and thus designers must be mindful and factor in sustainability when developing new devices. Emerging technologies may further enable more sustainable computing practices and cement the net-positive potential of TinyML.

**Acknowledgments.** We thank Carole-Jean Wu, Udit Gupta, David Brooks, Danilo Pau, and Paul Whatmough for their guidance and support.

---

f  https://www.tinyml.org/ and https://tinymledu.org/

**Shvetank Prakash** is a Ph.D. candidate in the John A. Paulson School of Engineering and Applied Sciences at Harvard University, Cambridge, MA, USA.

**Matthew Stewart** is a postdoctoral researcher at Harvard University, Cambridge, MA, USA.

**Colby Banbury** is a Ph.D. candidate in the John A. Paulson School of Engineering and Applied Sciences at Harvard University, Cambridge, MA, USA.

**Mark Mazumder** is a Ph.D. candidate in the John A. Paulson School of Engineering and Applied Sciences at Harvard University, Cambridge, MA, USA.

**Pete Warden** is CEO of Useful Sensors, Mountain View, CA, USA, and a Ph.D. candidate at Stanford University, Stanford, CA, USA.

**Brian Plancher** is an assistant professor of computer science at Barnard College, Columbia University, New York, NY, USA.

**Vijay Janapa Reddi** is John L. Loeb Associate Professor of Engineering and Applied Science at Harvard University, Cambridge, MA, USA.